\documentclass[conference]{IEEEtran}
\IEEEoverridecommandlockouts
\usepackage{cite}
\usepackage{amsmath,amssymb,amsfonts}
\usepackage{algorithmic}
\usepackage{graphicx}
\usepackage{textcomp}
\usepackage{xcolor}
\usepackage{multirow}
\usepackage{caption}
\usepackage{booktabs}
\usepackage{url} 
\usepackage[misc]{ifsym} 
\def\BibTeX{{\rm B\kern-.05em{\sc i\kern-.025em b}\kern-.08em
    T\kern-.1667em\lower.7ex\hbox{E}\kern-.125emX}}
\begin{document}

\title{GAInS: Gradient Anomaly-aware Biomedical Instance Segmentation\\}

\DeclareRobustCommand*{\IEEEauthorrefmark}[1]{%
  \raisebox{0pt}[0pt][0pt]{\textsuperscript{\footnotesize #1}}%
}

\author{
    \IEEEauthorblockN{
        Runsheng Liu\IEEEauthorrefmark{1}\textsuperscript{*}, 
        Hao Jiang\IEEEauthorrefmark{1}\textsuperscript{*},
        Yanning Zhou\IEEEauthorrefmark{2},
        Huangjing Lin\IEEEauthorrefmark{3}, 
        Liansheng Wang\IEEEauthorrefmark{4}, and 
        Hao Chen\IEEEauthorrefmark{1, 5, 6, \Letter}
    }
    \thanks{\textsuperscript{\Letter} Corresponding author: Hao Chen. Email: jhc@cse.ust.hk}
    \thanks{\textsuperscript{*} Equal contribution}

    \IEEEauthorblockA{
        \IEEEauthorrefmark{1} Department of Computer Science and Engineering, \\The Hong Kong University of Science and Technology, Hong Kong, China\\
        \IEEEauthorrefmark{2} Tencent AI Lab, Shenzhen, China \\
        \IEEEauthorrefmark{3} Imsight AI Research Lab, Shenzhen, China \\
        \IEEEauthorrefmark{4} Department of Computer Science,  Xiamen University, Xiamen, China \\
        \IEEEauthorrefmark{5} Department of Chemical and Biological Engineering, \\ The Hong Kong University of Science and Technology, Hong Kong, China\\
        \IEEEauthorrefmark{6} HKUST Shenzhen-Hong Kong Collaborative Innovation Research Institute, Futian, Shenzhen, China
    }
}

\maketitle

\begin{abstract}
Instance segmentation plays a vital role in the morphological quantification of biomedical entities such as tissues and cells, enabling precise identification and delineation of different structures. Current methods often address the challenges of touching, overlapping or crossing instances through individual modeling, while neglecting the intrinsic interrelation between these conditions.
In this work, we propose a \textbf{G}radient \textbf{A}nomaly-aware Biomedical \textbf{In}stance \textbf{S}egmentation approach  (\textbf{GAInS}), which leverages instance gradient information to perceive local gradient anomaly regions, thus modeling the spatial relationship between instances and refining local region segmentation. Specifically, GAInS is firstly built on a Gradient Anomaly Mapping Module (GAMM), which encodes the radial fields of instances through window sliding to obtain instance gradient anomaly maps. To efficiently refine boundaries and regions with gradient anomaly attention, we propose an Adaptive Local Refinement Module (ALRM) with a gradient anomaly-aware loss function.
Extensive comparisons and ablation experiments in three biomedical scenarios demonstrate that our proposed GAInS outperforms other state-of-the-art (SOTA) instance segmentation methods. The code is available at ~\url{https://github.com/DeepGAInS/GAInS}.
\end{abstract}


\section{Introduction}
\begin{figure*}[h!]
    \centering
    \centerline{\includegraphics[width=0.75\textwidth]{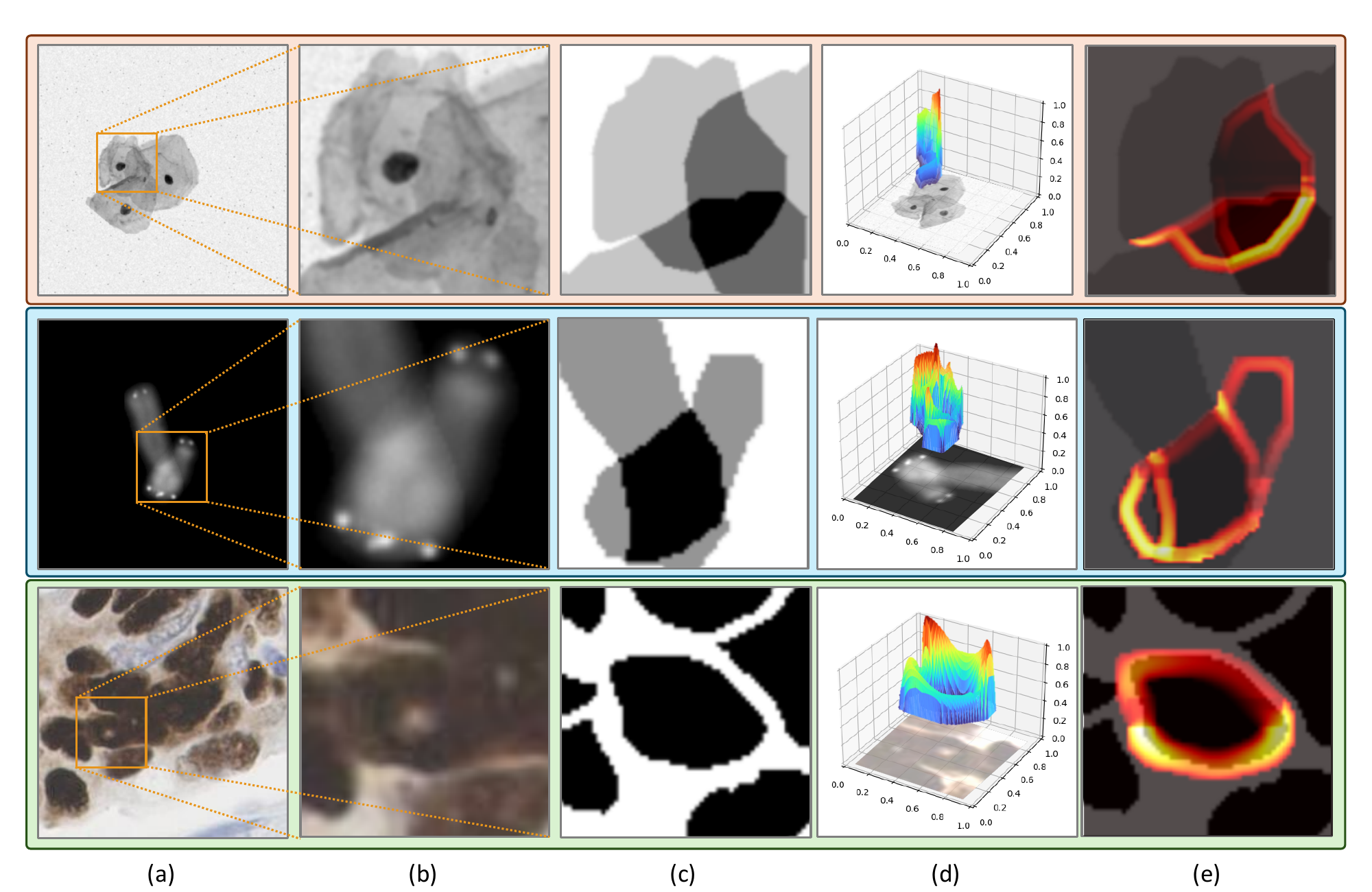}}
    \caption{The schematic illustration of biomedical instance segmentation: (a) medical images with \textbf{overlapping cells, crossing chromosomes, and touching nuclei}; (b) zoomed insets of local regions; (c) the corresponding ground truth masks; (d) gradient anomaly fields; (e) instance-level gradient anomaly maps.}
    \label{fig1}
    \end{figure*}

Instance segmentation is a fundamental task of biomedical image analysis, involving precise identification and delineation of biomedical objects or anatomical structures, such as chromosome, cell, gland, etc \cite{jiang2020geometry,chen2017dcan,fan2024dacseg}. This technique provides valuable supports for clinical decision-making by enabling subsequent qualification and assessment \cite{cheng2021robust}. For example, in the field of cytology, instance segmentation serves for identifying and quantifying abnormal cells for cancer screening \cite{jiang2022deep}. As shown in Fig. \ref{fig1}(a) and (b), biomedical instances often densely cluster, which causes touching, overlapping and crossing issues, raising challenges for instance segmentation.

Specifically, this issue could be attributed to dimensional compression caused by imaging, instances adhering and flattening during specimen preparation \cite{nayar2015bethesda}. Variations in instance arrangement and clustering further intensify the complexities of instance segmentation in the biomedical domain \cite{chen2016deep,li2023medshapenet,zhu2021hybrid,zhou2019irnet,zhou2020deep}. 

These challenges are compounded by the intricate interactions among instances. For example, interference of adjacent pixels leads to mis-classification and ambiguous segmentation boundaries. Besides, instance identification is problematic due to local interweaving of instances since it affects the model's perception of the instance integrity and morphology \cite{jiang2022deep,minaee2021image}. 

Modern instance segmentation approaches, which typically follow the detect-then-segment scheme \cite{he2017mask,minaee2021image,huang2019mask}, often struggle with touching and overlapping issues, leading to inaccuracies in instance detection and segmentation.
Previous methods address these challenges with application-specific solutions. For instance, DoNet \cite{jiang2023donet} is used for overlapping cytology segmentation, while auxiliary information like topology (TopoSeg \cite{he2023toposeg}) and direction information (HoverNet \cite{graham2019hover}, CDNet \cite{he2021cdnet}) are utilized in nuclei segmentation from pathology images.
However, these solutions are too tailored for specific scenarios. Simultaneous occurrence of touching and overlapping instances lacks a unified framework for comprehensive solutions. 

To establish a unified and general modeling approach for crossing, touching and overlapping instances, we propose utilizing gradient fields, which encapsulate information on distance and direction as effective guidance to model spatial information. By leveraging gradient fields, we intend to enhance the model's capability of capturing and representing regions that present instance touching and overlapping. We propose a \textbf{G}radient \textbf{A}nomaly-aware Biomedical \textbf{In}stance \textbf{S}egmentation approach (\textbf{GAInS}) involving Gradient Anomaly Mapping Module (\textbf{GAMM}) and Adaptive Local Refinement Module (\textbf{ALRM}). Specifically, Gradient Anomaly stands for the scenarios of local directional anomaly when the same pixel has multiple gradient directions, or there is a conflict of the gradient direction between adjacent pixels. The proposed GAInS starts from the typical two-stage baseline, namely Mask R-CNN \cite{he2017mask}, then follows an instance gradient anomaly modeling approach in GAMM. It maps ground truth (GT) (Fig. \ref{fig1}(c)) into gradient anomaly maps by a vector field transformation. 

Upon this, we introduce a gradient branch to learn the abnormal gradient field (shown in Fig. \ref{fig1}(d)) from local regions where instances \textbf{C}ross, \textbf{T}ouch and \textbf{O}verlap with each other (referred as \textbf{CTO} regions).
To further integrate gradient information into model learning for better instance separation, we introduce ALRM with a Gradient Anomaly-aware Mask Refinement Loss to attentively refine CTO regions by reweighing the mask loss according to Gradient Anomaly Maps. 
Ultimately, they altogether equip our model with the refinement capability in CTO regions, and improve final detection and segmentation performances.

\section{Methodology}
\begin{figure*}[t!]
    \centering
    \centerline{\includegraphics[width=1\textwidth]{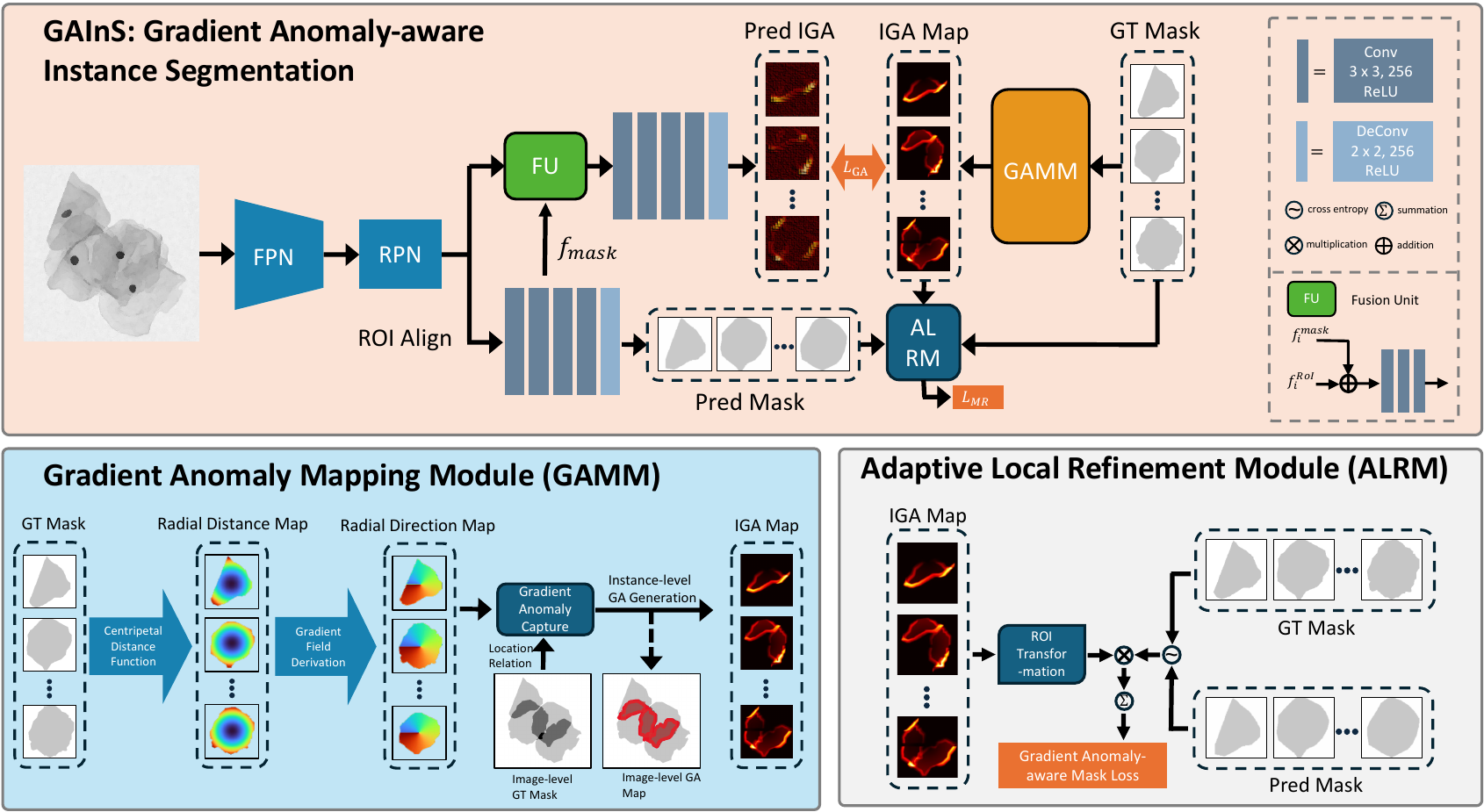}}
    \caption{Overview of the proposed GAInS. The network contains two main prediction branches: (1) gradient branch with GAMM for gradient anomaly learning and gradient anomaly field mapping; (2) mask branch with ALRM to adaptively refine CTO regions, such as the heavily overlapping regions of the cervical cells illustrated.}
    \label{fig2}
    \end{figure*}

Our GAInS, as provided in Fig. \ref{fig2}, adopts a two-stage proposal-based instance segmentation paradigm \cite{he2017mask}. It consists of backbone with Region Proposal Network (RPN) for instance candidate proposal. Then, the Region of Interest (RoI) features $f_{roi}$ are fed into gradient branch and mask branch for simultaneous predictions: semantic mask $\hat{M^{m}}$ and gradient anomaly map $\hat{M^{g}}$, which can efficiently capture the gradient information of instances. 

Specifically, the GT masks $M^{m}$, via the Gradient Anomaly Mapping Module (GAMM), generate gradient anomaly maps $M^{g}$ for each instance (Sec. \ref{sec2-2}). The gradient anomaly maps serve as the  GT of gradient branch prediction $\hat{M^{g}}$,  which is supervised by Gradient Anomaly Mask Loss $\mathcal{L}_{GA}$. This encourages the model to directly represent the gradient information. To further improve segmentation performance in CTO regions, we build ALRM in Sec. \ref{sec2-3} to adaptively refine instance pixels with gradient anomaly attention.
 
 The overall loss function of our framework can be formulated as,

 \begin{equation}
 \mathcal{L} = \mathcal{L}_{GA} + \mathcal{L}_{MR} + \mathcal{L}_{reg} + \mathcal{L}_{cls} + \mathcal{L}_{rpn},
  \label{eq:eq1}
\end{equation}
  where $\mathcal{L}_{GA}$ and $\mathcal{L}_{MR}$ denote the Gradient Anomaly Mask Loss and the Gradient Anomaly-aware Mask Refinement loss, $\mathcal{L}_{reg}$ is smooth-L1 regression loss and $\mathcal{L}_{cls}$ is cross-entropy classification loss, $\mathcal{L}_{rpn}$ contains losses for RPN supervision.

\subsection{Gradient Anomaly Modeling}\label{sec2-1}

In order to separate different instances with CTO regions, we first design a gradient branch to learn gradient information, and then construct gradient anomaly maps as the supervision. 

\begin{figure*}[t!]
    \centering
    \centerline{\includegraphics[width=0.85\textwidth]{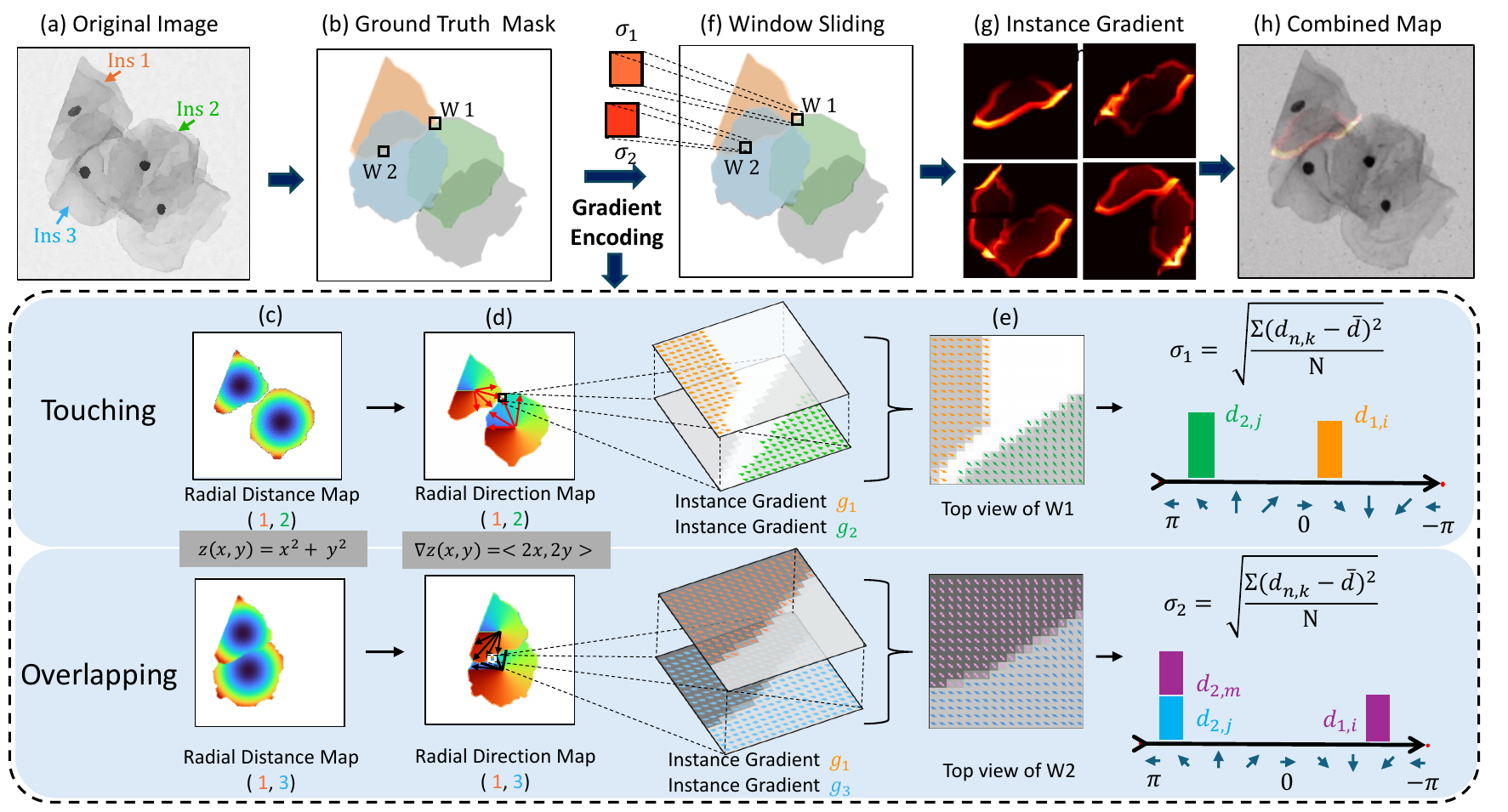}}
    \caption{The workflow of gradient anomaly map generation. (a) original images with (b) GT masks $M^{m}$ of interacting instances (\textcolor{orange}{$Ins1$}, \textcolor{green}{$Ins2$}, \textcolor{blue}{$Ins3$}); (c) radial distance map $R_{s}$; (d) radial direction map $R_{r}$; (e) top view $M_{W}$; (f) window sliding with deviation assignment; (g) gradient anomaly map $M^{g}$, and (h) combined image.}
    \label{fig3}
    \end{figure*}

As illustrated in Fig. \ref{fig3}, we define two types of gradient fields within each instance's mask, namely radial distance $ Rs = \{r_{1}^{s},...r_{i}^{s}\}$ and radial gradient $ Rr = \{r_{1}^{r},...r_{i}^{r}\}$, to capture the local instance's displacement and gradient anomaly.
We notice that these gradient fields could produce local directional anomaly in CTO regions. The abnormality occurs when the same pixel has multiple gradient directions, or there is a conflict of the gradient direction between adjacent pixels (\textcolor{green}{$g_{1}$} and \textcolor{orange}{$g_{2}$} in Fig. \ref{fig3}), which we call gradient anomaly. To highlight all CTO regions for each instance, we use sliding windows ($W_{1}$ and $W_{2}$ in Fig. \ref{fig3}) to enclose particular localities. The Gradient Anomaly Maps $M^{g}$ are then formed by computing the gradient anomaly within the windows and assigning values proportional to the response of gradient anomaly to each pixel. The technical details are described as follows. 

The first step of computing of the gradient anomaly map for each instance is to locate the geometric center points $(\bar{x}_i, \bar{y}_i)$ via their GT masks $M^{m}$ (Fig. \ref{fig3}(b)), followed by the establishment of a three-dimensional coordinate system. 
Then, a three-dimensional parabolic function, 
\begin{equation}
z(x_i,y_i) = (x_i-\bar{x}_i)^2+(y_i-\bar{y}_i)^2,
  \label{eq:eq2}
\end{equation}
 is implemented to generate a radial distance map $R_{s}$, which can be seen in Fig. \ref{fig3}(c). 

In the second step, we calculate the gradient of $z$,
\begin{equation}
\nabla z(x_i,y_i) = <2(x_i-\bar{x}_i),2(y_i-\bar{y}_i)>,
\label{eq:eq3}
\end{equation}
ensuring that the gradient direction of each pixel aligns with the radial direction. These maps are referred as radial direction maps $R_{r}$ (Fig. \ref{fig3}(d)), containing $r_{i}^{r}$ for each instance. 

The third step involves conducting pixel-wise measurements to assess the interaction between gradient fields of the instances.
As the sliding window $W$ traverses inside CTO regions, we collect statistics on the pixel direction angle $d_{i, k} = arctan(\frac{y_i}{x_i})$ in the multi-layer gradient fields within the window (Fig. \ref{fig3}(e)), where $i$ and $k$ denote instance and pixel.
Subsequently, we compute the standard deviation, 
\begin{equation}
\sigma = \sqrt{\frac{\Sigma(d_{i,k}-\bar{d})^2}{N}},
\label{eq:eq4}
\end{equation}
 which serves as a gradient anomaly value assigner for all pixels in the window (Fig. \ref{fig3}(f)). We follow this process, sliding the window and assigning $\sigma$ values to pixels within it, yielding $M^g$. Normalizing and multiplying $M^g$ by a factor $f_{GA}$ can adjust the final value of the map.

Leveraging on $M^g$ computed from previous steps, we propose a gradient branch to efficiently learn gradient information. As shown in Fig. \ref{fig2}, this branch consists of four convolutional layers, and it takes the fusion of RoI features and mask semantic features as inputs.
The learning process is supervised by the Gradient Anomaly Mask Loss, 
\begin{equation}
L_{GA} = \sum_{i=1}^{N}\frac{1}{n}(M_{i}^{g} - \hat{{M}_{i}^{g})}^2,
\label{eq:eq5}
\end{equation}
 where $M_{i}^{g}$ is the gradient anomaly map, $\hat{{M}_{i}^{g}}$ is the corresponding prediction, $n$ is the total number of pixels and $N$ is the total number of prediction maps.

\subsection{Adaptive Local Refinement Module}\label{sec2-2}
Beyond enabling the model to assimilate the characteristics of $M^g$ generated by GAMM, we introduce ALRM with a Mask Refinement Loss $\mathcal{L}_{MR}$. Making advantage of the fact that each pixel is assigned its own degree of error-proneness in $M^g$ after GAMM, we refine the pixel-level prediction loss to increment the penalty imposed on CTO regions.

To elaborate, we adopt the vanilla cross entropy loss $L_{mask}$ for semantic supervision. $L_{mask}$ is then subjected to an pixel-wise multiplication with $M^{g}$, the Mask Refinement Loss $L_{MR}$ is presented as,
\begin{equation}
L_{MR} = M^{g} \odot L_{mask} \
= M^{g} \odot  \frac{1}{K} \sum_{k=1}^K \frac{1}{N_k} \sum_{i=1}^{N_k} \mathcal{L}_{c e}\left(\hat{M}_{i}^{m}, M_{i}^{m}\right),
\label{eq:eq6}
\end{equation}
 where $K$ is the number of classes, $N_k$ denotes the number of samples of the k-th class, $M_{i}^{m}$ denotes the GT masks and $\hat{M}_{i}^{m}$ denotes the prediction. Note that $\mathcal{L}_{ce}$ is an operation to calculate cross entropy value for each pixel in a map without summing them up. $L_{MR}$ is obtained after the reweighing and the summing up all the values.

GAMM assigns gradient anomaly value to each pixel according to the surrounding gradient field interactions and every pixel represents an extent of error-proneness. Therefore, pixel-wise refinement enables the mask semantic loss to be gradient anomaly-aware, and accurately represent the incurred loss, eventually improve the segmentation performance.

\subsection{Special Cases of Gradient Anomaly Generation}\label{sec2-3}

The intense gradient anomaly in the internal intersection region, where no edges are involved in window should be weakened by another statistical method, to highlight the more critical information of the overlapping edge. As the sliding window W traverses inside intersection regions, we additionally apply pixel-wise subtraction between layers and subsequently we compute the standard deviation value to obtain the result. This special treatment benefits certain datasets such as UOUC (see section {III}).

\section{Experiments and Results}
\begin{table*}[!t]
    \centering
    \small
    \renewcommand{\arraystretch}{1.2}
    \caption{Comparison results of GAInS with respect to different Window Sizes (ws) and GA factors ($f_{GA}$) in GAMM. The \textbf{bold} and \textcolor{blue}{blue} are the first and second best results. }
    \setlength{\tabcolsep}{6mm}
    \begin{tabular}{c|cc|cc|cc|cc}
    \toprule[1pt]
        \multirow{2}{*}{Attributes} & \multicolumn{2}{c|}{mAP↑} & \multicolumn{2}{c|}{AJI↑} & \multicolumn{2}{c|}{Dice↑} & \multicolumn{2}{c}{ F1↑} \\ \cline{2-9}
        ~ & Cell & Ave. & Cell & Ave. & Cell & Ave. & Cell & Ave. \\
    \midrule
        ws = 3 & \textcolor{blue}{62.03} & \textcolor{blue}{62.72} & 73.60 & \textcolor{blue}{76.37} & \textbf{92.19} & \textbf{91.34} & 86.37 & 91.29 \\ 
        ws = 5 & \textbf{64.16} & \textbf{63.71} & \textbf{75.02} & \textbf{76.82} & \textbf{92.19} & \textcolor{blue}{91.30} & \textbf{89.05} & \textcolor{blue}{92.62} \\ 
        ws = 7 & 61.20 & 62.05 & \textcolor{blue}{73.74} & 74.84 & \textcolor{blue}{92.08} & 91.18 & \textcolor{blue}{88.94} & \textbf{92.68} \\ 
        ws = 9 & 59.59 & 60.67 & 72.41 & 73.42 & 91.81 & 90.85 & 87.38 & 91.59 \\ 
        ws = 12 & 60.96 & 61.72 & 72.88 & 72.73 & 91.60 & 90.90 & 86.93 & 90.99 \\ \hline
        $f_{GA}$ = 0.4 & 61.37 & 62.43 & 73.32 & \textcolor{blue}{75.26} & 91.85 & \textcolor{blue}{91.21} & 88.64 & \textcolor{blue}{92.57} \\ 
        $f_{GA}$ = 0.5 & \textbf{64.16} & \textbf{63.71} & \textbf{75.02} & \textbf{76.82} & \textbf{92.19} & \textbf{91.30} & \textbf{89.05} & \textbf{92.62} \\ 
        $f_{GA}$ = 0.75 & \textcolor{blue}{63.06} & \textcolor{blue}{62.44} & \textcolor{blue}{74.41} & 73.59 & \textcolor{blue}{91.92} & 90.99 & \textcolor{blue}{88.74} & 91.22 \\
    \bottomrule[1pt]
    \end{tabular}
    \label{tab2}
    \vspace{-0.05in}
\end{table*}

\subsection{Dataset and Experiments Settings}

\noindent\textbf{Dataset.} We evaluate GAInS on three medical instance segmentation datasets, including 1) ISBI2014\footnote{ISBI2014: \url{https://cs.adelaide.edu.au/~carneiro/isbi14_challenge/}} \cite{ISBI2014origin} which contains 945 cytology images with more than 3700 instances; 2) Kaggle2018\footnote{Kaggle2018: \url{https://www.kaggle.com/datasets/gangadhar/nuclei-segmentation-in-microscope-cell-images}}, focusing on the ‘cluster nuclei’ subset, with 885 instances; 3) UCOC\footnote{UCOC: \url{https://www.kaggle.com/datasets/jeanpat/ultrasmall-coco-125-overlapping-chromosomes}} for chromosome segmentation, containing 125 images with 250 chromosomes. We follow the offical setting of ISBI2014 to use 45, 90, 810 images for training, validation and testing to evaluate our method. The division of train, validation, and test sets in Kaggle2018 and UCOC is done by a ratio of 6:2:2 and a random seed. 

\noindent\textbf{Evaluation Metrics.}
The performance is evaluated using metrics including Jaccard index (AJI) \cite{kumar2017dataset}, average Dice coefficient (Dice), F1-score (F1), and mean of Average Precision (mAP).

\noindent\textbf{Implementation Details.}
We utilize the Mask R-CNN in Detectron2 \cite{wu2019detectron2} as the baseline model. Except for Mask R-CNN (R101), ResNet-50-based FPN network is the default for all experiments. In training stage, We adopt SGD with 0.9 momentum as the optimizer, set 0.015 initial learning rate and 20k training iterations to ensure the training convergence. We utilize 24GB Nvidia GeForce RTX3090 GPUs to conduct experiments. 

\subsection{Ablation Studies and Analysis}

\noindent\textbf{Effectiveness of Instance Comparison.}
First, we perform ablation studies using ISBI2014 dataset to investigate the effects of each component proposed in GAInS. The comparison results can be seen in Table. \ref{tab1}. 
By adding ALRM, we observe 3.99\%, 0.85\% increases in the average mAP and Dice. It suggests that only refining the mask loss properly could result in better performance, since the the mask losses of pixels in CTO regions are logically amplified, making them matters more in the process of backward propagation of the network. 
Meanwhile, only adding gradient anomaly learning (GAL) brings 2.86\%, 0.65\% increases on mAP and Dice. We can infer that equipping the model with the knowledge of gradient anomaly, the features of challenging regions to segment are well learnt, leading to higher scores of performance metrics.
Moreover, combining these two components increases the average mAP, AJI, Dice by 4.12\%, 0.61\%, 0.83\%, respectively. The perception of gradient information on CTO regions through GAL and the pixel-level adaptive refinement through ALRM altogether make GAInS more well-performed. 

\begin{table}[!h]
    \centering
    \small
    \renewcommand{\arraystretch}{1.2}
    \caption{Effect of each proposed module on ISBI2014. The \textbf{bold} and \textcolor{blue}{blue} denote the first and second best results.}
    \begin{tabular}{ccc|c|c|c|c}
    \toprule[1pt]
        Baseline & GAL & ALRM & mAP↑ & AJI↑ & Dice↑ & F1↑ \\ 
    \midrule
        \checkmark & ~ & ~ & 59.59 & \textcolor{blue}{76.21} & 90.47 & \textbf{93.14} \\ 
        \checkmark & ~ & \checkmark & \textcolor{blue}{63.58} & 74.58 & \textbf{91.32} & 92.29 \\ 
        \checkmark & \checkmark & ~ & 62.45 & 75.62 & 91.12 & 91.55 \\ 
        \checkmark & \checkmark & \checkmark & \textbf{63.71} & \textbf{76.82} & \textcolor{blue}{91.30} & \textcolor{blue}{92.61} \\
    \bottomrule[1pt]
    \end{tabular}
    \label{tab1}
    \vspace{-0.1in}
\end{table}

\noindent\textbf{Design Choice for GAMM.} We provide comparisons on ISBI2014 dataset to demonstrate the effectiveness of different attribute values: 1) Window Size, ranging from 3 to 12, determines the size of the local scope for computing gradient anomaly within it. Being too small or too large can both result in inferior performance. 2) GA factor, namely $f_{GA}$, ranging from 0.4 to 0.75, determines the value of $M^g$ so that influences the relative importance of distinct pixels when conducting the pixel-level refinement process in ALRM. Specifically, we use $f_{GA}$ to multiply the normalized gradient anomaly maps then yield $M^g$. Both window size and GA factor play crucial roles in the determination of the total loss, positively influence the weight updates in backward propagation, and thus contributing to the high performance of detection and segmentation. For datasets with different features of instances, the two attributes should be set properly to adapt the different key challenges in each datasets. Shown as an example, Table. \ref{tab2} indicates that we obtain the best results on ISBI2014 when window size is 5 and $f_{GA}$ is 0.5. To conclude, both window size and $f_{GA}$ should be chosen appropriately to fit different datasets.

\subsection{Comparison with SOTA methods}

\begin{table*}[t!]
    \vspace{0.1in}
    \centering
    \small
    \renewcommand{\arraystretch}{1.2}
    \caption{Comparison with other methods on ISBI2014, UCOC and Kaggle2018. The \textbf{bold} denotes the best results.}
    \label{tab4}
    \setlength{\tabcolsep}{7mm}
    \begin{tabular}{c|cc|cc|cc}
    \toprule[1pt]
        \multirow{2}{*}{Methods} & \multicolumn{2}{c|}{ISBI2014} & \multicolumn{2}{c|}{UCOC} & \multicolumn{2}{c}{Kaggle2018} \\ \cline{2-7}
        ~ & mAP↑ & AJI↑ & mAP↑ & AJI↑ & mAP↑ & AJI↑ \\
    \midrule
        Mask R-CNN (R50) \cite{he2017mask} & 59.59 & 76.21 & 72.33 & 81.29 & 38.75 & 54.79 \\
        Mask R-CNN (R101) \cite{he2017mask} & 62.97 & 75.15 & 73.71 & 80.87 & 37.13 & 52.80 \\ 
        Mask Scoring R-CNN \cite{huang2019mask} & 60.03 & 71.78 & 70.31 & 83.21 & 37.32 & 52.81 \\
        PISA \cite{cao2020prime} & 60.84 & 74.56 & 73.24 & 81.73 & 38.09 & 51.73 \\
        Cascade R-CNN \cite{cai2018cascade} & 63.40 & 52.21 & 73.66 & 81.09 & 40.33 & 54.09 \\
        CondIns \cite{tian2020conditional} & 49.46 & 59.79 & 50.57 & 66.88 & 38.41 & 46.50 \\
        HTC \cite{chen2019hybrid} & 62.57 & 35.95 & 70.32 & 83.69 & 37.73 & 53.48 \\
        Pointrend \cite{kirillov2020pointrend} & 62.07 & 69.45 & 71.14 & 31.48 & 37.95 & 52.63 \\
        Occlusion R-CNN \cite{follmann2019learning} &62.35& 78.64 &67.30&83.52&35.85&51.81 \\
        DoNet \cite{jiang2023donet} & 63.43 & \textbf{79.88} & 70.97 & 82.87 & 37.83 & 53.96 \\
        FastInst \cite{he2023fastinst} & 61.28 & 71.66 & 72.74 & 80.12 & 37.02 & 50.93 \\ 
        \hline
        GAInS & \textbf{63.71} & 76.82 & 73.94 & 83.59 & \textbf{40.63} & 53.62 \\
        GAInS (R101) & 61.39 & 77.66 & \textbf{74.48} & \textbf{85.87} & 39.79 & \textbf{55.30} \\
    \bottomrule[1pt]
    \end{tabular}
\end{table*}

\noindent\textbf{Quantitative Results.}
As shown in Table. \ref{tab4}, we quantitatively compare the instance segmentation results from the ISBI2014, UCOC and Kaggle2018 with the SOTA methods (Mask R-CNN \cite{he2017mask} with ResNet50 and ResNet101, Mask Scoring R-CNN \cite{huang2019mask}, PISA \cite{cao2020prime}, Cascade R-CNN \cite{cai2018cascade}, CondIns \cite{tian2020conditional}, HTC \cite{chen2019hybrid}, Pointrend \cite{kirillov2020pointrend}, Occlusion R-CNN \cite{follmann2019learning}, DoNet \cite{jiang2023donet} and FastInst \cite{he2023fastinst}). The proposed GAInS achieves the highest scores of mAP on ISBI2014, UCOC and Kaggle2018. Specifically, our proposed GAInS gains 0.31\% improvement on mAP with the best of others on the ISBI dataset, as well as 0.23\%, 0.30\% improvements on mAP with the best of others on UCOC dataset and Kaggle2018 dataset respectively. Meanwhile, GAInS achieves the second highest AJI on UCOC among all the SOTA methods. The significant increment of mAP and other metrics on all datasets show great advantage of GAInS on instance segmentation tasks. Moreover, compared with FastInst, a query-based model for instance segmentation, our proposed method achieves 5.17\%, 3.47\% and 2.69\% higher AJI score in ISBI2014, UCOC, and Kaggle2018. The constant improvement shows the rather better capability of GAInS to tackle with crossing, touching and overlapping issues than other latest works. Additionally, we implement the ResNet101-based GAInS and obtain further improvements on three datasets respectively. In particular, the ResNet101-based GAInS achieves 0.77\% mAP, 2.18\% AJI higher than the best of other SOTA methods on UOUC, and 0.51\% AJI higher than the best of others on Kaggle2018. 

\begin{figure*}[h!]
    \centering
    \centerline{\includegraphics[width=\textwidth]{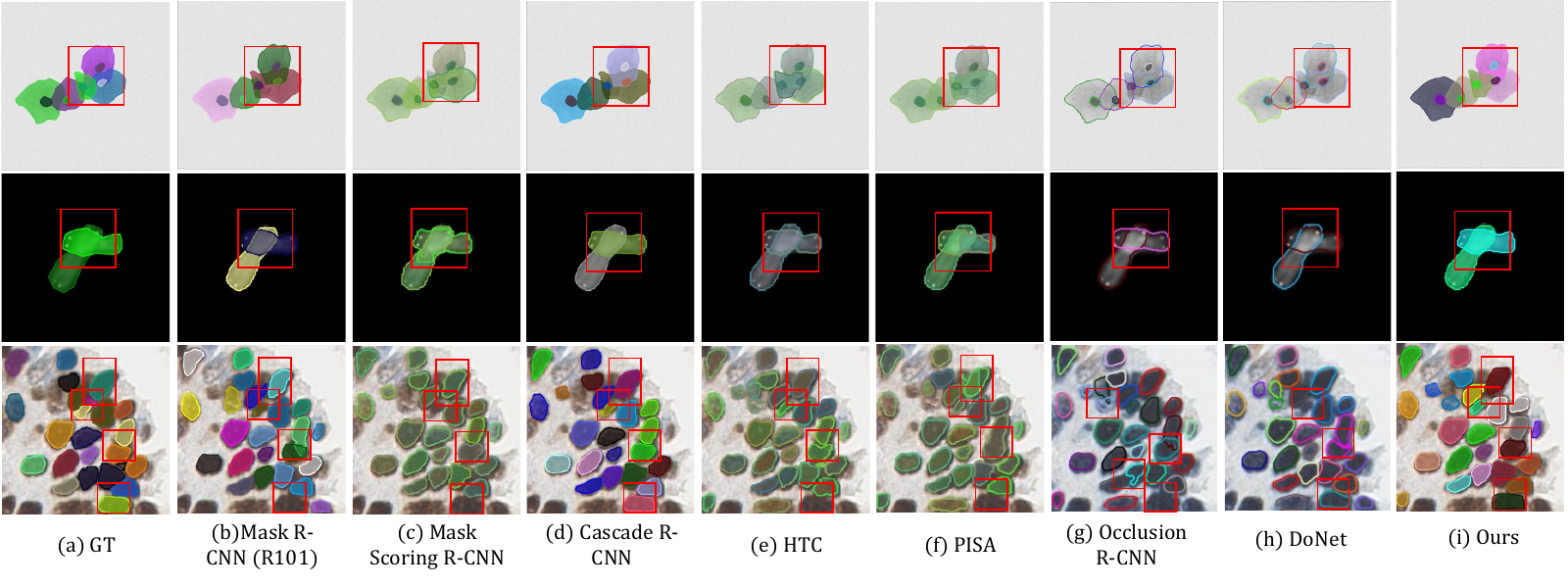}}
    \caption{Qualitative results of our GAInS and other SOTA methods. The red boxes highlight the CTO regions of cell, chromosome and nucleus.}
    \label{fig4}
\end{figure*}

\noindent\textbf{General Applicability on Different Key Challenges.}
Comparing with DoNet, which focuses on overlapping challenge such as ISBI2014, GAInS obtains significantly higher performance in UCOC and Kaggle2018, which have heavy crossing and touching problems. DoNet utilizes the information of intersection and complement area of overlapping cells, but this method does not benefit in the case of touching or crossing. In contrast, GAInS gives attention to touching contours and crossing or overlapping areas by sliding window enclosure, making use of information of all adjacent cells. The result shows the general applicability of GAInS for crossing, touching and overlapping challenges. Similarly, Occlusion R-CNN deals with overlapping issues but lacks general applicability to solve crossing and touching issues, thus it has less metric scores especially on UOUC and Kaggle2018. Overall, GAInS obtains higher performances in all the three datasets, resolving the three key issues and showing its general applicability.

\noindent\textbf{Qualitative Results.}
We compare segmentation results from the three datasets with several state-of-the-art instance segmentation models. Fig. \ref{fig4} shows representative challenging cases such as overlapping cell, crossing chromosome and touching nucleus. These difficult cases lead to missing detection and wrong segmentation in most of the models except for GAInS. This result shows that GAInS performs well when dealing with various cases of difficulties. For example, the fully overlapped cervical cell (first line of Fig. \ref{fig4}) is correctly detected only by GAInS. Meanwhile, the segmentation performance of GAInS shows high accuracy, especially when segmenting the CTO regions (shown in red boxes of Fig. \ref{fig4}), where instance boundaries are interfered by surrounding instances. Overall, our proposed GAInS successfully identifies and segments those error-prone regions, improving upon other methods.

\section{Conclusion}
To establish a unified and general approach for crossing, touching, and overlapping instances in biomedical instance segmentation, we propose GAInS, a novel approach that leverages instance gradient information, involving a Gradient Anomaly Mapping Module and an Adaptive Local Refinement Module. GAInS utilizes interactions of gradient fields to capture inherent gradient anomaly, so it gains perception and imposes refinements on the error-prone regions and boundaries. Experiments show that GAInS effectively handles instance segmentation tasks featuring different key issues, improving upon existing methods. Further exploration could be done to verify the adaptability of GAInS on more biomedical scenarios, such as multiple myeloma plasma cytology \cite{gupta2023segpc} and dental X-ray cephalometric images \cite{bolelli2023tooth}.

\section{Acknowledgments}
This work was supported by the National Natural Science Foundation of China (No. 62202403), Hong Kong Innovation and Technology Fund (Project No. PRP/034/22FX), Shenzhen Science and Technology Innovation Committee Fund (Project No. KCXFZ20230731094059008) and the Project of Hetao Shenzhen-Hong Kong Science and Technology Innovation Cooperation Zone (HZQB-KCZYB-2020083).

{\small
\bibliographystyle{IEEEtranS}
\bibliography{ref}
}
\end{document}